\documentclass{article}
\usepackage{ijcai18}

\usepackage{times}
\usepackage{xcolor}
\usepackage{soul}
\usepackage[utf8]{inputenc}
\usepackage[small]{caption}
\usepackage{graphicx}
\usepackage{amsmath}
\usepackage{algorithm}
\usepackage[export]{adjustbox}

\title{A Theoretical Investigation of Graph Degree as an Unsupervised Normality Measure}

\author{{Caglar Aytekin, Francesco Cricri, Lixin Fan, Emre Aksu}\\ 
Nokia Technologies, Tampere, Finland  \\
Corresponding Author's Email: caglar.aytekin@nokia.com}

\begin{document}

\maketitle

\begin{abstract}
For a graph representation of a dataset, a straightforward normality measure for a sample can be its graph degree. Considering a weighted graph, degree of a sample is the sum of the corresponding row's values in a similarity matrix. The measure is intuitive given the abnormal samples are usually rare and they are dissimilar to the rest of the data. In order to have an in-depth theoretical understanding, in this manuscript, we investigate the graph degree in spectral graph clustering based and kernel based point of views and draw connections to a recent kernel method for the two sample problem. We show that our analyses guide us to choose fully-connected graphs whose edge weights are calculated via universal kernels. We show that a simple graph degree based unsupervised anomaly detection method with the above properties, achieves higher accuracy compared to other unsupervised anomaly detection methods on average over 10 widely used datasets. We also provide an extensive analysis on the effect of the kernel parameter on the method's accuracy.

\end{abstract}


%

\section{Introduction}
\label{Intro}

Unsupervised anomaly detection aims to detect samples that deviate from the norm of the data, with no annotation available. 
This makes the problem challenging as the only available information is the internal structure of a dataset. 
Graph-based methods in general make use of the internal data structure, therefore these methods can be used for unsupervised anomaly detection as well.
The dataset can be represented by a graph where each node corresponds to a sample and each edge describes a connection between two samples. 
In a weighted graph, the weight of each edge indicates the similarity between two samples.
Then, the weighted graph degree of a node is the sum of the edge weights that are incident to that node. 
This measure can be considered as a intuitive representation for normality, since in a dataset, the populated dense clusters are strong indicators of normality and samples in these clusters usually have high degree.

The above intuition is not clearly observed from graph degree formulation (sum of the values in a row of similarity matrix).
Therefore, in this manuscript, we explicitly formulate graph degree in several points of view so that the formulations clearly show that the measure is fit for normality score. Our contributions are as follows.


\begin{figure}[!t]
\includegraphics[width=0.5\textwidth,center]{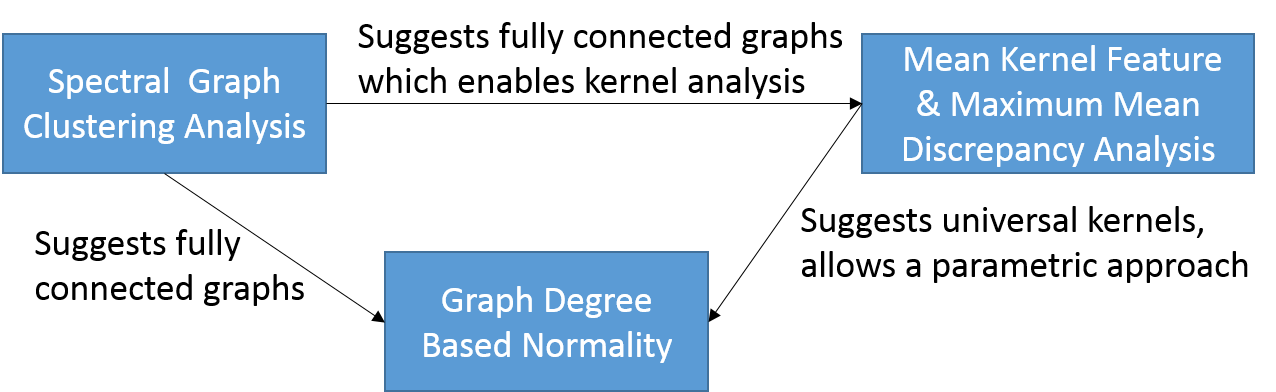}
\caption{Analyses of graph degree based normality and how the analyses guide the method.}
\label{theory}
\end{figure}

\begin{itemize}

\item We provide a spectral graph clustering based analysis of graph degree and show how the analysis guides us to use fully connected graphs.
\item We provide a kernel mean feature based and a maximum mean discrepancy based analysis of graph degree for fully connected graphs and show how the analysis guides us to use universal kernels.
\item Adopting fully connected graphs with a particular choice of a universal kernel, we evaluate an anomaly detection method based on graph degree and show its higher performance on average over 10 datasets, compared to other unsupervised anomaly detection methods.
\item We show that the kernel analysis allows a parametric approach to deal with challenging anomaly cases and we provide an extensive analysis on the effect of the parameter on the method's accuracy.

\end{itemize}


\section{Unsupervised Anomaly Detection: A Brief Review}
\label{RelWork}
An extensive review of unsupervised anomaly detection methods can be found in \cite{review}. 
Here we follow the same categorization to briefly review unsupervised anomaly detection algorithms by two main categories: $k$ nearest neighbor ($k$-nn) based and clustering based methods. 

For a sample $i$, represented by the feature $f_i$, the $k$-nn based anomaly detection \cite{knn} defines the following as an anomaly measure.
\begin{equation}
\tilde{d}_{i}^{(k)}=\frac{1}{|knn(i)|}\sum_{j\in{knn(i)}}\sqrt{(f_i-f_j)^T(f_i-f_j)},
\end{equation}
where $knn(i)$ is the set of $k$ nearest neighbors of sample $i$. 
This simple measure is observed to accurately highlight global anomalies, i.e. anomalies that are far away from all normal classes. However, it is not as accurate in detecting local anomalies, i.e. anomalies that are significantly closer to a particular normal cluster compared to others, yet still they form an anomaly case for that particular cluster. 
In order to address the drawback of $k$-nn baseline, variants with local density measures have been introduced. 
These methods LOF \cite{LOF}, COF \cite{COF} exploit the measure in Eq. \ref{lof}. LOF uses a direct Euclidean neighborhood definition, whereas COF replaces Euclidean distance with shortest path distance.
\begin{equation}
\frac{1}{|knn(i)|}\sum_{j\in{knn(i)}}\frac{\tilde{d}_{i}^{(k)}}{\tilde{d}_{j}^{(k)}}\label{lof}
\end{equation}
For further robustness to samples that are at intersection of two normal clusters, INFLO enhances the neighborhood of $i$ with including the reverse $k$-nn neighbors of $i$ as well as its $k$-nn neighbors. 
Here, reverse $k$-nn corresponds to the samples which have X in their $k$-nn. Other research in this category of studies include assigning an anomaly probability instead of a raw measure (LoOP \cite{LOOP}) and an automatic selection framework for $k$ of $k$-nn (LOCI \cite{LOCI}, aLOCI \cite{LOCI}).

Another category in unsupervised anomaly detection is clustering-based methods. 
CBLOF \cite{CBLOF} first clusters the data and classifies the clusters as large or small by a heuristic. 
Next, the distance $d^{(cl)}_i$ between a sample feature $f_i$ to its nearest cluster centroid $^ic$ multiplied with number of samples in the cluster is treated as an anomaly measure. 
In order to prevent anomalies that form small clusters, for small clusters this distance is taken between the data sample and the closest large cluster centroid. 
There is also a variant of CBLOF without the scaling with number of data samples in a cluster (UCBLOF \cite{review}). 
As an extension to CBLOF, LDCOF \cite{LDCOF} exploits the following ratio as an anomaly measure:
\begin{equation}
\frac{d^{(cl)}_i}{\frac{1}{|^ic|}\sum_j{d_j^{(cl)}}}. 
\end{equation}
This is done in order to handle local anomalies as well. 
There is also another study CMGOS \cite{CMGOS} which follows LDCOF, but uses Mahalanobis distance instead of Euclidean. 
All clustering based methods generally make use of k-means due to its linear complexity, however this makes the methods very dependent on the choice of k.

Other works which do not fall into above main categories are histogram based outlier score (HBOS) \cite{HBOS}, 1-class support vector machine (1-class SVM) \cite{oSVM} and robust Principle Component Analysis (rPCA) \cite{rPCA} based anomaly detection methods.

\section{Graph Degree as a Normality Score}
\label{PropMet}

In this section we will analyze the graph degree from three points of view: a) spectral graph based clustering and b) kernel feature mean and, c) maximum mean discrepancy points of view.

\subsection{A Spectral graph clustering analysis}
A basic spectral graph clustering method \cite{maxassoc} aims to find a cluster that maximizes the average intra-cluster similarity.
Consider a cluster indicator vector $\pmb{x}$, entries of which takes 1 if a sample belongs to the cluster and 0 otherwise.
Then the average intra-similarity of the cluster for can be written as follows:

\begin{equation}
 \frac{\textbf{x}^TW\textbf{x}}{\textbf{x}^T\textbf{x}} \label{eqn1}
\end{equation}


In Eq. \ref{eqn1}, $W$ is a symmetric similarity (or affinity) matrix where $w_{ij}$ indicates the similarity between nodes $i$ and $j$. 
In Eq. \ref{eqn1}, numerator is the sum of similarities of every sample pair in the class and denominator is the total number of samples in that class.

The optimal solution maximizing Eq. \ref{eqn1} is intractable, therefore one can relax the vector $\pmb{x}$ to have real values. Then, optimal solution is the eigenvector $\pmb{\psi}$ of $W$ with maximum eigenvalue due to Rayleigh Quotient \cite{maxassoc}. 
Other solutions can also be obtained with other eigenvectors, where the eigenvalue $\lambda_i$ indicates the wellness of solution (higher the better).
This is straightforward as $\lambda_i=\frac{\pmb{\psi_i}^TW\pmb{\psi_i}}{\pmb{\psi_i}^T\pmb{\psi}_i}$.

From an anomaly detection point of view, the clusters that are populated and dense can be considered to correspond to \emph{normal} clusters.
Combining such clusters would highlight normal data, therefore suppressing anomaly data.
The clusters with high intra-cluster similarity are dense clusters.
Since eigenvectors are soft indicators of all clusters and since eigenvalues are measures of denseness, a straightforward combination would be as follows.

\begin{equation}
\sum_{i=1} \lambda_i\pmb{\psi}_i. \label{eqn2}
\end{equation}

Although this combination might seem intuitive and it covers the denseness assumption (due to weighting with $\lambda_i$), it is not clear whether the combination also highlights populated clusters.
This is because the eigenvectors are only assumptions to binary cluster indicators and it is not analyzed if it makes sense to combine them blindly.
Next, we provide an analysis of eigenvector values and we provide extension to Eq. \ref{eqn2} in order to combine clusters by weighting both dense and populated clusters.

Assume for a cluster $c_i$ , the corresponding eigenvector takes the same value within the cluster and zero otherwise. 
Then this value will be $\sqrt{N_{c_i}}$, since the eigenvector is $L_2$ normalized. 
Note here that $N_{c_i}$ is the number of samples in cluster $c_i$.
Then, the combination in Eq. \ref{eqn2} can be considered as a weighted sum of binary cluster indicators where the weight is $\lambda_i \sqrt{N_{c_i}}$.

An intuitive combination of binary cluster indicators would include the $N_{c_i}$ term directly instead of its square-root, since a populated cluster is a direct indicator of normality.
Therefore, we multiply Eq. \ref{eqn2} with $\sqrt{N_{c_i}}$.
It can be easily seen that $\pmb{\psi}_i^T\textbf{1}=\sqrt{N_{c_i}}$.
Therefore, adding this multiplicative term to Eq. \ref{eqn2}, we end up with Eq. \ref{eqn3}.

\begin{equation}
\pmb{\nu}= \sum_{i=1} \lambda_i\pmb{\psi}_i\pmb{\psi}_i^T\textbf{1}. \label{eqn3}
\end{equation}

\begin{figure*}[!t]
\includegraphics[width=0.85\textwidth,center]{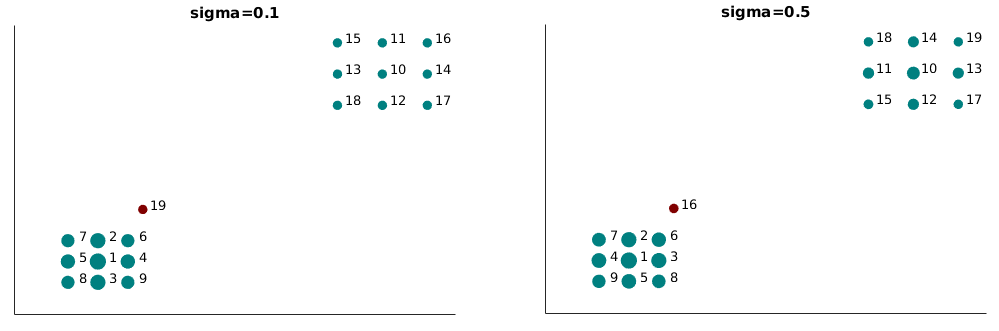}
\caption{An example on the effect of $\sigma$ on finding anomalies. Number of rankings of descending normality score are shown, e.g. samples with lower rank (large number) is less normal.}
\label{whykernelfig}
\end{figure*}

Next, we show that $\pmb{\nu} $ is equivalent to the graph degree.

Since the eigenvectors are orthonormal due to symmetric $W$, they form a basis and any vector can be written as a linear combination of the eigenvectors where weights are determined by the dot product of the vector with eigenvectors. Thus, one can rewrite the vector \textbf{1} as a combination of eigenvectors as follows:

\begin{equation}
\textbf{1}=\sum_{i=1} \pmb{\psi}_i\pmb{\psi}_i^T\textbf{1}. \label{eqn4}
\end{equation}
Multiplying Eq. \ref{eqn4} with $W$ from both sides we can obtain \ref{eqn6}.
\begin{equation}
W\textbf{1}=\sum_{i=1} W\pmb{\psi}_i\pmb{\psi}_i^T\textbf{1}, \label{eqn6}
\end{equation}
From the eigenvector relation $W\pmb{\psi}_i=\lambda_i\pmb{\psi}_i$, one can rewrite Eq. \ref{eqn6} as follows. 

\begin{equation}
W\textbf{1}=\sum_{i=1} \lambda_i\pmb{\psi}_i\pmb{\psi}_i^T\textbf{1} \label{eqn7}
\end{equation}

Right hand sides of Eq. \ref{eqn7} and Eq. \ref{eqn3} are equivalent, thus the following holds:
\begin{equation}
\label{eqn5}
\pmb{\nu}=W\textbf{1}.
\end{equation} 

Here, we have shown the equivalence of the graph degree and a spectral graph clustering based normality score.
The normality score highlights populated and dense clusters.
This analysis is a theoretical justification of the soundness of graph degree as a normality score from a clustering perspective.

In order to fully exploit the assumptions in this analysis, one needs to have a fully connected graph. 
This is explained as follows.
Constraining the connectivity of the graph to a rule enforces some entries of the similarity matrix to be zero.
This might incorrectly assign zero similarity to samples in the same cluster.
This is not desired since then the nominator of Eq. \ref{eqn1} would not exactly be the sum of all possible pair similarities within a cluster.

Thus, the spectral graph clustering based approach guides us to have a fully connected graph.

\subsection{A Mean Kernel Feature Based Analysis}
\label{whykernel}

Here we investigate graph degree as a kernel based normality score and discuss this interpretation.
It should be noted that the analysis applies to fully connected graphs which have been favored by our spectral graph clustering based analysis above.

Let us assume that for the fully connected graph, the graph edge weights are defined by a kernel, i.e. $W=K$, where  $K_{ij}=\pmb{\phi}_i^T\pmb{\phi}_j$. Here $\pmb{\phi}_i$ is the kernel feature of a data sample $i$. Then, one can rewrite Eq. \ref{eqn5} as follows.

\begin{equation}
 \pmb{\nu}={\Phi}^T{\Phi}\textbf{1} ,\label{meankernel}
\end{equation}

In Eq. \ref{meankernel}, ${\Phi}$ is a matrix containing kernel features for all samples. It can be easily realized that $\Phi\textbf{1}=N\tilde{\pmb{\phi}}$, where $\tilde{\pmb{\phi}}$ is the empirical estimation of the kernel feature mean and $N$ is the total number of samples in the dataset.

For a probability distribution $\textit{p}$ which the data is sampled from, the mean embedding of $\textit{p}$ into the Hilbert space is given by the kernel feature mean $\tilde{\pmb{\phi}}$ \cite{MMD}. 
The mapping is injective, i.e. each $\textit{p}$ is mapped to a unique element $\mu_\textit{p}$, for universal kernels including the RBF kernel \cite{MMD}.

Thus, for universal kernels, kernel mean feature is a descriptor of the probability distribution where the data is drawn and no other distribution can be described by that mean feature. Hence, the similarity of a sample to the mean in kernel space is then a sound normality score. We provide further analysis about this and relation to maximum mean discrepancy in Section \ref{MMD}.


In the light of the above analysis, we consider a universal kernel -RBF kernel- given in Eq. \ref{rbf}.
\begin{equation}
K_{ij}=e^{-\frac{(f_i-f_j)^2}{2\sigma^2}} \label{rbf}
\end{equation}
In Eq. \ref{rbf}, $f_i$ corresponds to feature related to sample $i$. Then, defining the normality as the degree of the kernel matrix $K$, allows us to have a parametric approach where the behavior of the method varies with varying $\sigma$.

This is especially important, since a parametric approach would help to overcome some limitations of the method.
By using the RBF kernel, a non-linear warping of the Euclidean space with parameter $\sigma$ is possible and thus with the right $\sigma$, some challenging anomaly types can be handled.
One toy example is illustrated in Fig. \ref{whykernelfig}.
The normal classes are generated by 2 densely connected clusters and there exists a local anomaly near the first cluster indicated by red. 
The second normal cluster is slightly more loosely connected compared to the first normal cluster.
Sample features are simply Cartesian coordinates.
The figure illustrates two selections of $\sigma$ and next to each element, that element's rank is written where the ranking is based on descending normality score.
The radii of the balls that represent elements are scaled accordingly to their normality score, larger elements mean larger normality score.
As it can be  observed, while selecting $\sigma=0.1$ can correctly push the local anomaly to the end of the rank, selecting $\sigma=0.5$ assigns less normality scores to some elements of the second normal cluster than what is assigned to abnormal sample.
It is observed in this toy example that an optimal warping of the space would help to detect challenging anomaly types as well.

\subsection{Relation to Maximum Mean Discrepancy}
\label{MMD}

In this section, we provide a supplementary analysis that also encourages using universal kernels. In particular, we show how the graph degree is related to a kernel based method for two sample problem \cite{MMD}. 
The two sample problem is defined as follows.

Let $p$ and $q$ be two distributions and $X$, $Y$ observations drawn from $p$ and $q$ respectively.
Then the problem is to determine if $p=q$.
A related measure to the similarity of two distributions is the maximum mean discrepancy ($MMD$). Let $F$ be a class of functions $f$, then $MMD$ and its empirical approximation are defined as follows.

\begin{equation}
\begin{aligned}
MMD(F,p,q)=\sup_{f \in F}(E_{x \sim p}[f(x)]-E_{y \sim q}[f(y)])  \\
MMD(F,X,Y)=\sup_{f \in F}(\frac{1}{m}\sum_{i=1}^{m} f(x_i)-\frac{1}{n}\sum_{i=1}^{n} f(y_i))
\end{aligned} \label{eqnMMD}
\end{equation}

Recently, a kernel method for the two sample problem was proposed \cite{MMD}.
The main result of this paper is as follows.
Let $F$ be a unit ball in reproducing kernel Hilbert space $H$ defined on metric space with associated kernel $k(.,.)$. Then,

\begin{equation}
MMD(F,p,q)=0 \iff p=q \label{mainres}
\end{equation}

An unbiased test for squared $MMD$ is given as follows.

\begin{equation}
\begin{split}
MMD^2(F,p,q)=E_{x,x' \sim p}[k(x,x') + E_{y,y' \sim q}[k(y,y')] \\
-2E_{x \sim p, y \sim q}[k(x,y)]]  
\end{split}\label{unbiased}
\end{equation}

The corresponding empirical estimation can be obtained as follows.

\begin{equation}
\begin{split}
MMD^2(F,X,Y)=\frac{1}{m(m-1)} \sum_i^m \sum_{i \neq j}^m k(x_i, x_j) \\
+ \frac{1}{n(n-1)} \sum_i^n \sum_{i \neq j}^n k(y_i, y_j) - \frac{2}{mn} \sum_i^m \sum_j^n k(x_i, y_j)
\end{split}\label{unbiasedempiric}
\end{equation}

Next, we investigate $MMD^2(F,X,Y)$ for a special case where $Y:\{x_l\}$, i.e. a case where the dataset $Y$ contains of only one element and that element is also contained in dataset $X$. Let us assume that we have an RBF kernel, thus $k(x_l,x_l)=1$. Consider now that we form a graph out of $X$ and the affinity matrix is defined by $K$, where $K_{ij}=k(x_i,x_j)$. Let $d_i$ denote the degree of sample $x_i$ in this graph. Then we can rewrite the first term in in Eq. \ref{unbiased} as follows.

\begin{equation}
\begin{split}
\frac{1}{m(m-1)} \sum_i^m \sum_{i \neq j}^m k(x_i, x_j)=\frac{1}{m(m-1)} \sum_i^m (d_i-1) \\
=\frac{d_{avg}-1}{m-1},
\label{firstterm}
\end{split}
\end{equation}

where $d_{avg}$ is the average degree of a sample in the dataset $X$.

The second term in Eq. \ref{unbiased} reduces to 1 since dataset $Y$ contains only one sample, whereas the third term can be written as follows. 

\begin{equation}
\begin{split}
\frac{2}{mn} \sum_i^m \sum_j^n k(x_i, y_j) = \frac{2d_l}{m} \label{thirdterm}
\end{split}
\end{equation}

Note that $k(x_i, y_1)=k(x_i, x_l)$ since the only element of $Y$ is equal to $x_l$. Then, combining terms, one can rewrite Eq. \ref{unbiased} for our special datasets $X,Y$ as follows.

\begin{equation}
\begin{split}
MMD^2(F,X,Y)= \frac{d_{avg}-1}{m-1}+1-\frac{2d_l}{m} \label{degreeMMD}
\end{split}
\end{equation}

        \begin{table*}
    \centering
    \caption{Evaluation of Anomaly Detection Methods: \textbf{\textcolor{blue}{best}}, \textbf{\textcolor{red}{second}} and \textbf{\textcolor{magenta}{third}} methods are highlighted}
    \resizebox{\textwidth}{!}{
    \label{comparison}
    \begin{tabular}{|c|c|c|c|c|c|c|c|c|c|c|c|}
        \hline
         & bcancer & pen-g & pen-l & letter & speech & satellite & thyroid & shuttle & aloi & kdd99 & avg \\\hline
       knn & 0.9791 & \textbf{\textcolor{blue}{0.9872}} & 0.9837 & 0.8719 & 0.4966 & \textbf{\textcolor{blue}{0.9701}} & 0.5956 & 0.9424 & 0.6502 & 0.9747 & \textbf{\textcolor{red}{0.8452}} \\\hline
       kthnn & \textbf{\textcolor{magenta}{0.9807}} & \textbf{\textcolor{red}{0.9778}} & 0.9757 & 0.8268 & 0.4784 & \textbf{\textcolor{red}{0.9681}} & 0.5748 & 0.9434 & 0.6177 & 0.9796 & \textbf{\textcolor{magenta}{0.8323}} \\\hline
       lof&\textbf{\textcolor{red}{0.9816}}&0.8495&\textbf{\textcolor{blue}{0.9877}}&0.8673&0.5038&0.8147&0.647&0.5127&0.7563&0.5964&0.7517\\\hline
       lofub&0.9805&0.8541&\textbf{\textcolor{red}{0.9876}}&\textbf{\textcolor{magenta}{0.9019}}&\textbf{\textcolor{magenta}{0.5233}}&0.8425&0.6663&0.5182&\textbf{\textcolor{magenta}{0.7713}}&0.5774&0.7623\\\hline
       cof&0.9518&0.8695&0.9513&0.8336&0.5218&0.7491&0.6505&0.5257&\textbf{\textcolor{red}{0.7857}}&0.5548&0.7394\\\hline
       inflo&0.9642&0.7887&0.9817&0.8632&0.5017&0.8272&0.6542&0.493&0.7684&0.5524&0.7395 \\\hline
       loop&0.9725&0.7684&\textbf{\textcolor{magenta}{0.9851}}&\textbf{\textcolor{red}{0.9068}}&\textbf{\textcolor{red}{0.5347}}&0.7681&\textbf{\textcolor{magenta}{0.6893}}&0.5049&\textbf{\textcolor{blue}{0.7899}}&0.5749&0.7496\\\hline
       loci&0.9787&0.8877&NA&0.788&0.4979&NA&NA&NA&NA&NA&0.7881\\\hline
       aloci&0.8105&0.6889&0.8011&0.6208&0.4992&0.8324&0.6174&0.9474&0.5855&0.6552&0.7058\\\hline
       cblof&0.2983&0.319&0.6995&0.6792&0.5021&0.5539&0.5825&0.9037&0.5393&0.6589&0.5736\\\hline
       ucblof&0.9496&0.8721&0.9555&0.8192&0.4692&0.9627&0.5469&0.9716&0.5575&\textbf{\textcolor{red}{0.9964}}&0.8101\\\hline
       ldcof&0.7645&0.5948&0.9593&0.8107&0.4366&0.9522&0.5703&0.8076&0.5726&0.9873&0.7456\\\hline
       cmgosr&0.914&0.5693&0.9727&0.7711&0.5077&0.9054&0.4395&0.5425&0.5852&0.7265&0.6934\\\hline
       cmgosg&0.8992&0.6994&0.9449&0.8902&0.5081&0.9056&0.6587&0.5679&0.5855&0.9797&0.7639\\\hline
       cmgosm&0.9196&0.6265&0.9038&0.7848&NA&0.912&\textbf{\textcolor{red}{0.8014}}&0.6903&0.5547&0.9696&0.7959\\\hline
       hbos&\textbf{\textcolor{blue}{0.9827}}&0.7477&0.6798&0.6216&0.4708&0.9135&\textbf{\textcolor{blue}{0.915}}&\textbf{\textcolor{magenta}{0.9925}}&0.4757&\textbf{\textcolor{blue}{0.999}}&0.7798\\\hline
       rpca&0.9664&0.9375&0.7841&0.8095&0.5024&0.9461&0.6574&\textbf{\textcolor{blue}{0.9963}}&0.5621&0.7371&0.7899\\\hline
       osvm&0.9721&\textbf{\textcolor{magenta}{0.9512}}&0.9543&0.5195&0.465&0.9549&0.5316&0.9862&0.5319&0.9518&0.7819\\\hline
       nsvm&0.9581&0.8993&0.9236&0.7298&0.4649&0.943&0.5625&0.9848&0.5221&0.7945&0.7782\\\hline
       GDBA&0.9403&0.8998&0.9763&\textbf{\textcolor{blue}{0.9284}}&\textbf{\textcolor{blue}{0.6332}}&\textbf{\textcolor{magenta}{0.9679}}&0.6382&\textbf{\textcolor{red}{0.9944}}&0.5677&\textbf{\textcolor{magenta}{0.994}}&\textbf{\textcolor{blue}{0.8540}}\\\hline
    \end{tabular}}
\end{table*} 

Based on the measure in Eq. \ref{degreeMMD}, if we define an anomaly measure for a sample $x_l$ in a dataset $X$ as $MMD^2(F,X,Y)$ where $Y:{x_l}$, then the anomaly measure is inversely correlated with the graph degree. For a sorting based method, i.e. anomaly samples are determined according to their positions in a sorted data based on anomaly measure, then the graph degree is the only determining factor for anomaly measure. This is due to the constant value of the first two terms of Eq. \ref{degreeMMD}.

From a distribution discrepancy point of view, $MMD^2$ corresponds to the discrepancy between the data generating probability distribution $p$, from which the data $X$ is sampled from and a distribution $q$ which is defined as a Dirac delta function at sample $x_l$ , i.e. it is a deterministic process. In this sense, the similarity of a sample to the entire dataset is proportional with the the inverted $MMD^2$ and thus with graph degree.

However, inverted $MMD^2$ can indicate similarity to other datasets as well, since the mean embedding of two different probability distributions can be same. Only way to avoid this is to select universal kernels such that the mean embedding of probability distribution $p$ is injective. Therefore, the $MMD$ based analysis also suggests selecting universal kernels.

\section{Experimental Results}
\label{expres}
In this section, we evaluate the graph degree based anomaly detection in widely used anomaly detection datasets and compare it to the widely used unsupervised anomaly detection methods. For a fair comparison, we follow the datasets, methods and evaluation metrics in the benchmark paper \cite{review}. Next, we briefly describe these and our implementation details.

\subsection{Implementation Details}
We perform feature standardization as a preprocessing step, i.e. we make the mean of each feature zero and we make the standard deviation of each feature one across the dataset. 
Based on our theoretical analysis in spectral graph clustering analysis point-of-view, we choose to have a fully connected graph, where each node is connected to the other. 
Based on our theoretical analysis in kernel mean feature and maximum mean discrepancy point of views, due to its universal property, we use RBF kernel in Eq. \ref{rbf} to form the kernel matrix with an empirical selection of $\sigma=0.15$.
The effect of $\sigma$ on the performance will be analyzed in detail.
In order to eliminate the effect of data dimension, we always normalize the $(f_i-f_j)^2$ term in Eq. \ref{rbf} with the data dimension.
After the degree vector $\nu$ of $K$ is calculated as in Eq. \ref{eqn5}, the anomaly score is simply obtained by inverting $\nu$.

\subsection{Datasets}

We use Breast Cancer Wisconsin \cite{bcancer}, Pen-Based Recognition of Handwritten Text -Global and Local \cite{UCI}, Letter Recognition \cite{letter}, Speech Accent Data \cite{letter}, Landsat Satellite \cite{UCI}, Thyroid Disease \cite{UCI}, Statlog Shuttle \cite{UCI}, Object Images ALOI \cite{aloi} and KDD-Cup99 HTTP \cite{UCI} datasets. Due to limited space, we omit description of each dataset in detail and suggest the reader to refer to \cite{review} for detailed information about the datasets. 

\begin{table*}
    \centering
    \caption{Deviation of performance with varying $\sigma$ and performance at best $\sigma$. The cases where the method is \textbf{\textcolor{blue}{best}}, \textbf{\textcolor{red}{second}} and \textbf{\textcolor{magenta}{third}} are highlighted. }
    \label{analysis}
    \resizebox{\textwidth}{!}{
    \begin{tabular}{|c|c|c|c|c|c|c|c|c|c|c|}
        \hline
         & bcancer & pen-g & pen-l & letter & speech & satellite & thyroid & shuttle & aloi & kdd99  \\\hline
        GDBA&0.9448&0.9413&0.9808&0.9206&0.6006&0.9686&0.6206&0.9944&0.5602&0.994\\
        & $\pm0.0042$ & $\pm0.0522$ & $\pm0.0109$ & $\pm0.0037$ & $\pm0.0225$ & $\pm0.0129$ & $\pm0.0498$ & $\pm0.0117$ & $\pm0.0451$ & $\pm0.0001$  \\\hline
       GDBAbest&\textbf{\textcolor{blue}{0.9829}}&\textbf{\textcolor{red}{0.9842}}&0.9810&\textbf{\textcolor{blue}{0.9302}}&\textbf{\textcolor{blue}{0.6490}}&\textbf{\textcolor{red}{0.9689}}&\textbf{\textcolor{magenta}{0.7665}}&\textbf{\textcolor{red}{0.9946}}&0.7407&\textbf{\textcolor{magenta}{0.994}}\\
       (best $\sigma$) &(0.885)&(0.32)&(0.1950)&(0.11)&(0.035)&(0.1750)&(0.0250)&(0.1850)&(0.0050)&(0.19)\\\hline
    \end{tabular}}
\end{table*}

\subsection{Compared Methods}
The compared methods include (a) $k$-nn based methods : $k$-nn \cite{knn}, $k^{th}$-nn \cite{kthnn}, LOF \cite{LOF}, LOFUB \cite{review}, COF \cite{COF}, INFLO \cite{INFLO}, LoOP \cite{LOOP}, LOCI \cite{LOCI}, aLOCI \cite{LOCI}. (b) clustering based methods: CBLOF \cite{CBLOF}, uCBLOF, LDCOF \cite{LDCOF}, CMGOS-Red \cite{review}, CMGOS-Reg \cite{review}, CMGOS-MCD \cite{review} and (c) other methods: HBOS \cite{HBOS}, rPCA \cite{rPCA}, oc-SVM \cite{oSVM} and $\mu$-oc-SVM \cite{oSVM}.

\subsection{Evaluation Metrics}
An anomaly detection method often generates an anomaly score, not a hard classification result. Therefore, a common evaluation strategy in anomaly detection is to threshold this anomaly score and form a receiver operating curve where each point is the true positive and false positive rate of the anomaly detection result corresponding to a threshold. Then, the area under the curve (AUC) of RoC curve is used as an evaluation of the anomaly detection method \cite{review}. 

\subsection{Discussion}

We compare the AUC of the GDBA with other methods in the anomaly detection datasets. The results are given in Table \ref{comparison}. 
Note that the results denoted by NA correspond to methods that take too long to implement (more than 12 hours) for that dataset.
Table \ref{comparison} illustrates that the GDBA outperforms other methods on average. 
GDBA's performance on individual datasets however does not show a clear leading accuracy in each dataset individually.
Still, GDBA is in top three best performing methods in 5/10 datasets.
As each dataset is different from each other in structure, the optimal nonlinear warping of the Euclidean space , i.e. the optimal parameter $\sigma$, can be different for the best accuracy.

\subsubsection* {Effect of $\sigma$} 

\begin{figure}[!t]
\includegraphics[width=0.5\textwidth,left]{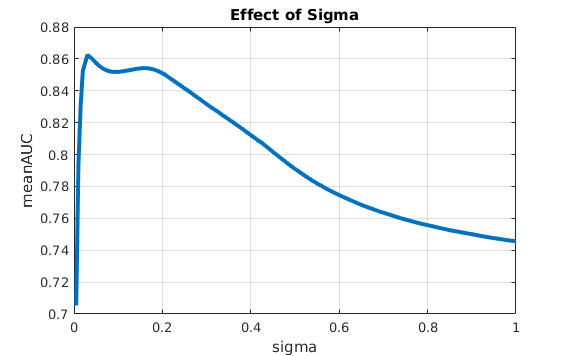}
\caption{Effect of $\sigma$: The average area under the ROC curve across all datasets with different $\sigma$.}
\label{fig0}
\end{figure}

In Fig. \ref{fig0}, we illustrate the effect of the parameter on the accuracy via observing the average performance (AUC) of the method across all datasets while changing $\sigma$. We observe a somewhat robust performance when $\sigma \in [0.02,0.2]$. The mean of the method's performance and the standard deviation (indicated with $\pm$) across this sigma interval is given in Table \ref{analysis}. 

In the same table, we also report the best AUC measure with the manually selected $\sigma$ in all datasets under the algorithm named GDBAbest. Note that the best sigma is chosen by a grid search across $\sigma \in [0.005,1]$. We observe that the potential of the method is high considering its leading performance on average by a good margin. Moreover, 8 out of 10 datasets, the method is in top three. This is of course if the hyper-parameter $\sigma$ is selected optimally. 

In our experiments, we could not find a direct correlation between $\sigma$ and data statistics such as feature dimensionality, number of samples, anomaly percentage, number of normal or abnormal classes etc. However, the selection of $\sigma$ is dependent on the type of anomalies that the dataset contains. For example, the datasets with local anomalies require small $\sigma$, this is due to the need to separate the local anomalies from the normal data such that they become global anomalies. Thyroid, aloi and speech are such datasets and the optimal performance is obtained at very low sigma. For datasets where a single normal class is present and anomalies are not local, the sigma is best selected high (see bcancer dataset). This is in order to cluster the normal data as much as possible. Since the anomalies are global they will not be merged with normal cluster even if $\sigma$ is high. However, in pen-g dataset, although there is one large normal class, the anomalies are not so far away from the normal cluster, so very large $\sigma$ does not help in this case, but a middle value is optimal. 

The optimal $\sigma$ is difficult to predict without using annotated validation set. In a semi-supervised anomaly detection setting, where we have the labels in a validation set, $\sigma$ can be optimized to maximize the performance in validation dataset. But this is not the case we are interested in this work, as we aim to have an unsupervised method.



\section{Conclusion}
\label{Conc}
We have analyzed the graph degree measure as a normality score in spectral graph clustering based point of view and mean kernel feature based point of view and introduced how it is related to maximum mean discrepancy. Our analyses verify the theoretical soundness of graph degree as a normality score, moreover they guide us to use fully connected graphs with universal kernels.
In our experiments, we have observed that a simple graph degree based anomaly detection with an empirical parameter selection outperforms all other methods on average in 10 datasets. We have also observed that optimal parameter selections would result in a much improved performance. We have investigated the performance dependency on this parameter $\sigma$ and optimal selection of the $\sigma$ parameter per dataset. We have observed that the method is somewhat robust to $\sigma$ in an interval. Finally, the automatic selection of $\sigma$ remains as an open problem as we have found that there is no direct correlation of $\sigma$ to any data statistics such as feature dimensionality, number of samples, or ratio of outliers.

%
%
%
%
%
%

\bibliographystyle{named.bst}
\bibliography{ijcai18.bib}

\end{document}